\documentclass{SCIS2026}
\usepackage{algorithm}
\usepackage{algorithmic}
\begin{document}
%\oa
%%%%%%%%%%%%%%%%%%%%%%%%%%%%%%%%%%%%%%%%%%%%%%%%%%%%%%%
%%% Authors do not modify the information below
%%% ×÷Õß²»ÐèÒªÐÞ¸Ä´Ë´¦ÐÅÏ¢
\ArticleType{RESEARCH PAPER}
%\SpecialTopic{}
\Year{2025}
\Month{January}
\Vol{68}
\No{1}
\DOI{}
\ArtNo{}
\ReceiveDate{}
\ReviseDate{}
\AcceptDate{}
\OnlineDate{}
\AuthorMark{}
\AuthorCitation{}
%%%%%%%%%%%%%%%%%%%%%%%%%%%%%%%%%%%%%%%%%%%%%%%%%%%%%%%

%%% title: ±êÌâ
%%%   \title{title}{title for citation}
\title{Wireless communication empowers online scheduling of partially-observable transportation multi-robot systems in a smart factory}{Title for citation}

%%% Corresponding author: Í¨ÐÅ×÷Õß
%%%   \author[number]{Full name}{{email@xxx.com}}
%%% General author: Ò»°ã×÷Õß
%%%   \author[number]{Full name}{}
%%% Equal Contribution: Í¬µÈ¹±Ï××÷Õß
%%%   \author[number\dag]{Full name}{}

\author[1]{Yaxin Liao}{}
\author[1]{Qimei Cui}{{cuiqimei@bupt.edu.com}}
\author[2]{Kwang-Cheng Chen}{}
\author[1]{Xiong Li}{} 
\author[1]{\\Jinlian Chen}{} 
\author[1]{Xiyu Zhao}{}
\author[1]{Xiaofeng Tao}{}
\author[3]{Ping Zhang}{}

%%% Authors' contribution. 
%\contributions{These authors contributed equally to this work.}

%%% Address.
%%%   \address[number]{Affiliation, City Postcode, Country}
\address[1]{National Engineering Research Center of Mobile Network Technologies, \\Beijing University of Posts and Telecommunications, Beijing 100876, China}
\address[2]{Department of Electrical Engineering, University of South Florida, Tampa, FL 33620 USA}
\address[3]{School of Information and Communication Engineering, Beijing University of Posts and Telecommunications, \\Beijing 100876, China}

%%% Abstract.
\abstract{Achieving agile and reconfigurable production flows in smart factories depends on online multi-robot task assignment (MRTA), which requires online collision-free and congestion-free route scheduling of transportation multi-robot systems (T-MRS), e.g., collaborative automatic guided vehicles (AGVs). Due to the real-time operational requirements and dynamic interactions between T-MRS and production MRS, online scheduling under partial observability in dynamic factory environments remains a significant and under-explored challenge. This paper proposes a novel communication-enabled online scheduling framework that explicitly couples wireless machine-to-machine (M2M) networking with route scheduling, enabling AGVs to exchange intention information, e.g., planned routes, to overcome partial observations and assist complex computation of online scheduling. Specifically, we determine intelligent AGVs' intention and sensor data as new M2M traffic and tailor the retransmission-free multi-link transmission networking to meet real-time operation demands. This scheduling-oriented networking is then integrated with a simulated annealing-based MRTA scheme and a congestion-aware A*-based route scheduling method. The integrated communication and scheduling scheme allows AGVs to dynamically adjust collision-free and congestion-free routes with reduced computational overhead. Numerical experiments shows the impacts from wireless communication on the performance of T-MRS and suggest that the proposed integrated communication and scheduling scheme significantly enhances scheduling efficiency compared to the local reasoning-based baseline, even under high AGV load conditions and limited channel resources. Moreover, the results reveal that the scheduling-oriented wireless M2M communication design fundamentally differs from human-to-human communications, implying new technological opportunities in a wireless networked smart factory.}


\maketitle

\section{Introduction}
Smart factories, an important vertical application in 6G mobile communications, are powered by reconfigurable cyber-physical multi-robot systems (MRS) with integrated artificial intelligence, computing, control, and networking technologies \cite{1}. To realize agile and reconfigurable production lines, smart factories rely on online multi-robot task assignment (MRTA) for production MRS \cite{2}. Furthermore, transportation MRS (T-MRS), e.g., collaborative autonomous guided vehicles (AGV), requires collision-free and congestion-free online scheduling \cite{1,niw}, which is significantly under-explored in literature.

\begin{figure}[ht]
	\centerline{\includegraphics[width=1\textwidth]{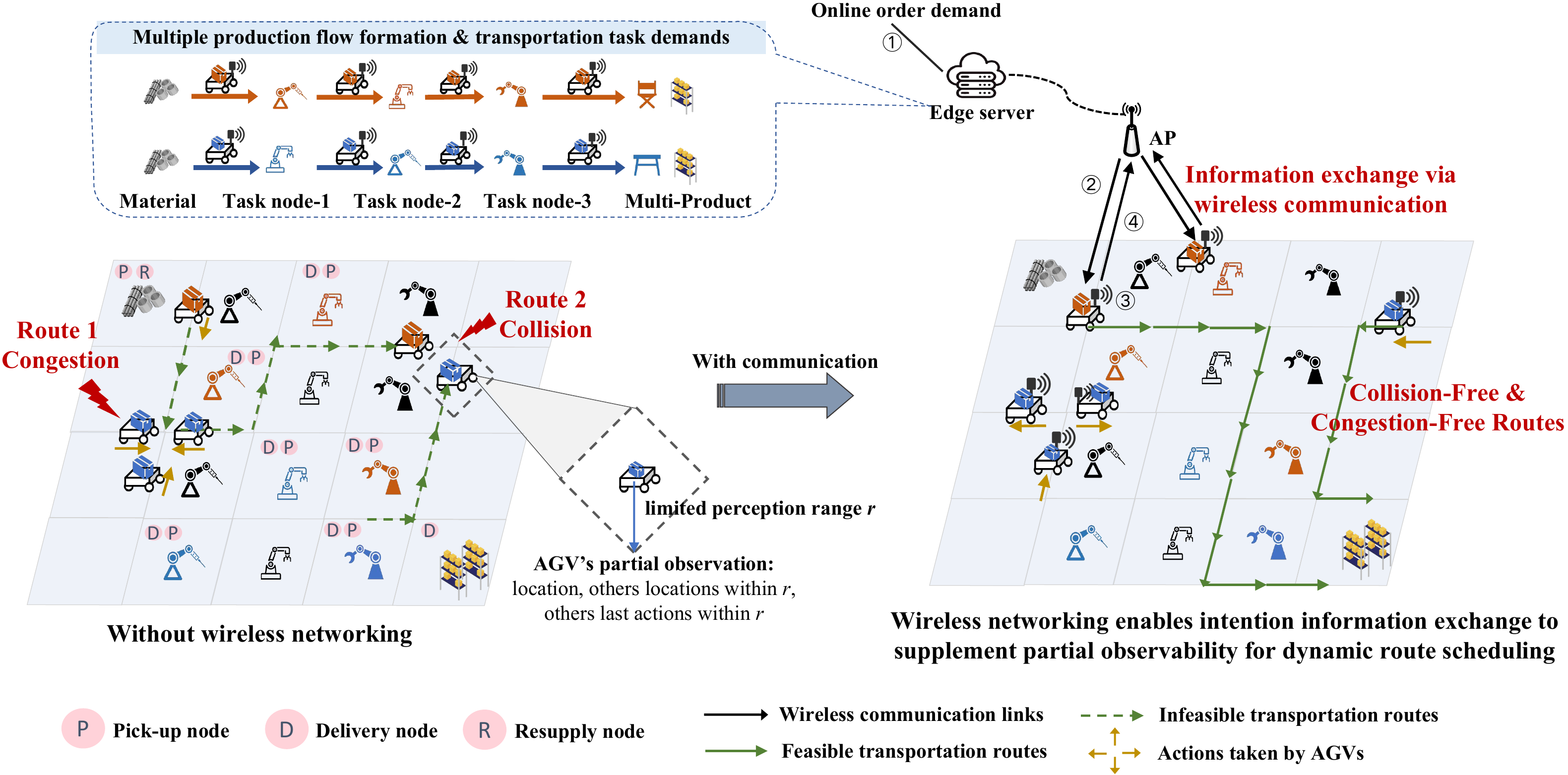}}
	\caption{ Online scheduling for MRS in a smart factory. The edge server performs MRTA to the production MRS and thus T-MRS. (\textcircled{1}$\rightarrow $\textcircled{2}). Without wireless networking, multi-AGV operating in a distributed manner with partial observations lead to collisions and congestion. With wireless networking, AGVs collaboratively execute pick-up and delivery tasks on time by real-time adjusting routes (\textcircled{3}), which significantly relies on AI operational intention and sensor data exchange via wireless networking (\textcircled{4}).} 
	\label{fig1}
\end{figure}

According to the online order demands, the edge server performs real-time MRTA and dynamically assigns production tasks to production MRS and dynamic transportation tasks to transportation MRS, as depicted in Fig.~\ref{fig1}. Therefore, to support agile and flexible production, T-MRS must accordingly execute route scheduling to ensure the on-time pickup and delivery of industrial materials for the production MRS. Due to the shared factory floor, collisions are possible when multiple AGVs simultaneously move to the same location. Moreover, congestion occurs when many AGVs traverse an intersection simultaneously in different directions. Given the latency and scalability bottlenecks of centralized control, each intelligent AGV must perform real-time actions in a distributed manner, but collectively avoid collisions and congestion.

However, online scheduling for T-MRS faces several challenges. AGVs operate in a distributed manner and can only observe nearby obstacles and other AGVs within their sensing radius due to limited onboard sensor capabilities \cite{task}. They are unaware of distant congestion, the planned motion intentions, and task information of others. This partial observability leads to coordination inefficiencies in distributed AI operations. Moreover, the highly dynamic factory environments impose real-time decision demands, requiring AGVs to adjust their routes dynamically based on incomplete local observations. 
In addition, realistic constraints further complicate scheduling, especially the dynamic interactions between production MRS and T-MRS. The completion of a production task triggers the start of its corresponding transportation task, which in turn affects when the subsequent production task can begin.

Wireless communication among mobile AGVs suggests a technological solution to the real-time coordination and scheduling challenges in dynamic and partially observable smart factory environments \cite{zhaojsac,zhaotvt}. By enabling AGVs to share locally partial observations and obtain globally fused information \cite{14}, communication enhances awareness of congestion and other AGVs' intention, thereby effectively reducing the uncertainty and supplementing partial observability. This allows AGVs to make more informed scheduling decisions and better adapt to dynamic factory conditions.
However, wireless machine-to-machine (M2M) communication differs fundamentally from the commonly used wireless human-to-human (H2H) communication. M2M communication is designed in response to the computational requirements of AI functionality by considering downstream tasks and data contents \cite{cui2025, TCOM1, TCOM2}, while H2H communication focuses on communication-related metrics optimization, overlooking the effects of wireless communication on subsequent task decision \cite{M2M}. 

Consequently, to support collision-free and congestion-free online scheduling, the design of wireless M2M networking should be oriented toward downstream scheduling tasks while considering its impact on the overall scheduling performance of the factory. By recognizing control and scheduling as computational goals, scheduling-oriented communication is a core issue within the integrated communication and computation (ICC) domain for 6G \cite{ICC, SCC2}. There is a pressing need to systematically explore tailored wireless networking to meet the computational demands of online scheduling and AI functionalities. To the best of our knowledge, no prior study has addressed this in the context of multi-robot smart factories.
\subsection{Related work}
Regarding the online scheduling for MRS in smart factories, some researchers study on dynamic flow-shop scheduling for production MRS \cite{PMRS,PMRS2,2}. But they implicitly assumed that AGVs perfectly complete transportation tasks and ignored the uncertainties of transportation in dynamic environment. Several works attempted to improve AGV scheduling. For example, \cite{tii} proposed a dynamic AGV scheduling model but neglected collision and congestion issues under partial observability. Heuristic methods \cite{5} addressed coupled task assignment and path planning but lacked proactive congestion avoidance. Other works studied collision-free navigation via reinforcement learning \cite{AGVsched}, or congestion-aware multi-robot planning \cite{congestion-aware1}. However, they overlooked the dynamic interaction between production MRS and T-MRS, which is critical in a smart factory. Although \cite{6} analyzed such interactions; however, it still did not address collision-free and congestion-free scheduling for AGVs. 
Due to the neglect of wireless networking design, the above studies usually make assumptions to simplify factory scheduling, which is not applicable and practical for online scheduling in partially observable and dynamic multi-robot smart factories.

Some studies explored wireless resource management for indoor robots \cite{h2h1,h2h2}, but primarily focused on optimizing communication performance alone, overlooking the MRS scheduling demands of the overall factory. Similarly, \cite{7} solely studied improving channel access rate by the topology control and channel allocation for the T-MRS. \cite{3} employed communication-based token exchange to coordinate T-MRS, but assumed ideal, error-free links and lacked explicit wireless networking design. Therefore, these works overlooked the scheduling demands of MRS and treated communication as an independent layer.

In summary, prior studies either (i) idealized the transport system and ignored collisions and congestion, (ii) considered AGV route scheduling but without heterogeneous production–transportation MRS interactions, or (iii) optimized wireless networking in isolation from online scheduling. 
None of these lines of work provides a systematic framework that integrates scheduling-oriented wireless networking to handle the production–transportation coupling and partial observability of dynamic smart factories.

\subsection{Contributions}

Building upon the identified gaps, the main contributions are summarized as follows:
\begin{itemize}
\item 
We propose a novel communication-enabled online scheduling framework for distributed T-MRS operating with online arrival tasks and dynamic interactions of heterogeneous MRS under partial observability. Unlike prior works that decouple communication from downstream tasks, our framework explicitly integrates wireless M2M networking into route scheduling by treating the AGV's planned intention as critical M2M traffic. By exchanging such intention information, we effectively supplement the local partial observations and assist complex computation of online scheduling in a highly dynamic factory environment.

\item We systematically tailor the wireless M2M networking to meet the real-time operational demands of online scheduling. Specifically, we design a retransmission-free multi-link transmission scheme, which exploits frequency diversity to mitigate cyber collisions and errors. This scheduling-oriented networking is then integrated with a simulated annealing-based MRTA scheme and a modified A*-based route scheduling method, allowing AGVs to dynamically adjust collision-free and congestion-free routes with reduced computational overhead.

\item By introducing real-time multi-link wireless transmission networking to achieve online scheduling, this paper shows the impacts from wireless communication on the performance of T-MRS and reveals several key findings: 1) Wireless communication among collaborative AGVs significantly enhances scheduling efficiency compared to the local reasoning-based baseline, even under high AGV load conditions. 2) Tailored real-time multi-link transmission networking effectively mitigates partial observability and enhances scheduling efficiency even with limited channel resources. 
3) The optimal solution in scheduling-oriented wireless M2M communication differs from the optimal solution in H2H communication, implying a fundamental difference and new technological opportunities in a wireless networked multi-robot smart factory.

\end{itemize}

The rest of this paper is organized as follows. Section~\uppercase\expandafter{\romannumeral2} formulates the online scheduling problem of T-MRS in a smart factory. Section~\uppercase\expandafter{\romannumeral3} provides the implementation of online scheduling. Section~\uppercase\expandafter{\romannumeral4} details the design of the wireless networking tailored to online scheduling.
Simulation results and new findings are presented in Section~\uppercase\expandafter{\romannumeral5}. Conclusions with future directions are provided in Section~\uppercase\expandafter{\romannumeral6}. The notations used in this paper are summarized in Table \ref{tab:t1}.

\begin{table*}
  \setlength{\extrarowheight}{1pt}
     \caption{\centering{Main Notations}}
    \centering 
    \scriptsize
        \scalebox{1}{
\begin{tabular}{|>{\centering\arraybackslash}m{0.1\columnwidth}|>{\centering\arraybackslash}m{0.3\columnwidth}|>{\centering\arraybackslash}m{0.1\columnwidth}|>{\centering\arraybackslash}m{0.3\columnwidth}|}
\hline
\textbf{Notation} & \textbf{Definition}  & \textbf{Notation} & \textbf{Definition} \\
\hline
$\mathcal{K}$ & The set of AGVs & $\mathcal{R}_k$,$\mathcal{P}_k$ & Task routes, and navigation routes of AGV $k$ \\
            \hline
$C$, $S$, $D$ &  The number of available channels, selected channels, and communication interval & $\textbf{B}$ & Decision matrix of MRTA \\
            \hline
$\widehat{o}_k^t$, $\tilde{o}_k^t$, $o_k^t$, $s^t$ & Local partial observation, additional observation, enhanced fusion observation, and global state & $\widehat{W}$ & Makespan based on local observations\\
            \hline
            $x_{m_+}$, $x_{m_-}$, $d_{m_+}^t$, $d_{m_-}^t$, $t_m^{ddl}$,${\mu}_{{\lambda}_m}$ & Task nodes for pick-up and delivery, quantity of production at task nodes, task's due time, task's priority & $M^t$ & the number of new tasks at time $t$ \\
            \hline
            $\widehat{AT}_{k,e}$ &  Reach time of AGV $k$ at task node & $\widehat{\textbf{A}}$ & Route decision matrix \\
            \hline
             $\widehat{WT}_{k,e}$ & waiting time of AGV $k$ at task node & $\widehat{PT}_{k,e}$ & preparation time at task node \\
            \hline
%\bottomrule
\end{tabular}
    }
    \label{tab:t1}
    
\end{table*}

\section{System model}
\subsection{Factory environment, definitions, and assumptions}
\textbf{The smart factory environment:} The smart factory transportation environment is characterized by an undirected graph, denoted as $G=(V,E)$, where $V$ is the set of all discrete locations in the factory and $E $ is the set of edges representing traversable paths between these locations. As illustrated in Fig.~\ref{fig1}, fixed-location production robots are deployed in columns by type and are located at the center of the grid cell. Each AGV in the T-MRS, denoted as $k\in \mathcal{K}=\{1,2,\dots K\}$, collaboratively transports industrial materials on time for production MRS based on partial observations to support sequential production processing. All AGVs share a fixed sensor range $r$, defining the maximum observation distance for any AGV.
Let $L\subseteq V$ and $U\subseteq V$ denote the sets of AGV-located nodes and resupply nodes, respectively.

\textbf{The definition of transportation tasks:} Let $M^t$ represent the total number of new transportation tasks, which arise from changes in production demands and can be adjusted at each reconfiguration timestep $t = t_1, t_2, \ldots$. The value of $M^t$ is determined by the number of production lines ${\lambda}^t$ and the number of corresponding tasks per production line ${\omega}^t$. Thus, $M^t = {\lambda}^t \cdot {\omega}^t ,t=t_1,t_2,\dots$. The new set of transportation tasks is denoted as $\mathcal{X}^{{t}}=\{x_1,x_2,\dots,x_{M^t}\}$. Each task consists of a sequence of two goal nodes: a pickup node and a delivery node. The pickup node is the partially completed products from an upstream production robot. The delivery node is the location of a downstream production robot that will continue the production process. To execute a task, an AGV must first visit the pickup node and then the delivery node within the allotted time. 

Formally, each transportation task $x_m \in \mathcal{X}^{{t}}$ is characterized by a tuple $(x_{m_+},x_{m_-},d_{m_+},d_{m_-},t_m^{ddl},{\mu}_{m})$. The pickup task node is denoted by $x_{m_+} \in P$, and the delivery task node by $x_{m_-} \in S$, where $P, S \subseteq V$. $d_{m_+}$ is the number of the partially completed products produced at pick-up node $x_{m_+}$. $d_{m_-}$ indicates the number of raw material demands at delivery node $x_{m_-}$. $t_m^{ddl}$ denotes the allotted time to accomplish task $x_m$. ${\mu}_{m}$ represents the priority of transportation task $x_m$ within the same production flow, where downstream tasks must wait for the completion of upstream tasks with higher priority \cite{6}. 

\textbf{The definition of AGV characteristics:} Each AGV is characterized by a tuple $(\widehat{o}_{k}^t,l_{k}^t,d_{k}^t)$. $\widehat{o}_{k}^t$ denotes the local partial observation of AGV $k$ at time slot $t$, which will be detailed later.
$l_k^t\in L$ is the location node of AGV $k$ and the state of the AGV is defined by its current location $l_k^t$. Let $K_{l_k^t}$ denote the number of AGVs located at $l_k^t$ at time $t$, which should be less than the threshold $ \kappa^*$ to avoid congestion. $d_k^t$ denotes the remaining raw materials currently carried by the AGV $k$. Thus, AGV $k$ needs to resupply raw materials at the node $u\in U$ when $d_k^t$ is less than the demand, $d_{m_-}$, of its assigned task. 
Assuming that the maximum payload of AGV is restricted by ${d}^*_k$. 
At each time slot $t$, AGV $k$ selects an action $\widehat{a}_k^t$ (move `north', `east', `south', `west', or `stay') to either move from its current vertex to an adjacent vertex or remain at the same vertex based on its partial observation. This action is also referred to as a ``navigation step''.

\textbf{The definition of partial observability:} 
Ideally, achieving collision-free and congestion-free scheduling requires each AGV to make decisions based on the full global state $s^t$. Crucially, this includes not only the real-time positions but also the future navigation intentions and dynamic task status of all other agents across the factory. However, due to limitations in sensor availability, quality, and perception range $r$, each AGV individually perceives only a subset of dynamic global state within its sensing range, denoted by $\widehat{o}_{k}^t$.
Formally, this partial observation is generated by an observation function $\mathcal{F}$ that maps the global state $s^t$ and perception range $r$ to the AGV's local perception \cite{5}, $\widehat{o}_{k}^t=\mathcal{F}(s^t,r), \forall k \in K$.

Specifically, $\widehat{o}_{k}^t$ consists of the AGV's own state and information about other nearby AGVs. To formalize ``nearby," we define the set of neighbors for AGV $k$ at time slot $t$ as all other AGVs within a sensing range $r$: $\mathcal{N}_k^t(r) = \left\{ j \in \mathcal{K} \setminus \{k\} \;\middle|\; \left\| l_k^t - l_j^t \right\|_1 \leq r \right\}$, where $\left\|\cdot\right\|_1$ denotes the Manhattan distance. Similarly, let $V(r) \subseteq V$ denote the set of nodes observable by AGV $k$ within its sensing range $r$.
Thus, the partial observation is a tuple containing its own location, the locations of its neighbors, and the last actions taken by those neighbors: $\widehat{o}_{k}^t \triangleq \{l_k^t,\{l_{j}^t\}_{j\in\mathcal{N}_k^t(r)},\{a_{j}^{t-1}\}_{j\in\mathcal{N}_k^t(r)}\}$. Consequently, $\widehat{o}_{k}^t$ inherently omits any information about locations and agents outside the AGV's sensing range, formalizing the concept of partial observability.

\textbf{The definition of task assignment route:} 
During each reconfiguration, new transportation tasks are assigned to AGVs. This assignment is formalized by a binary decision matrix $\mathbf{B}\in\{0,1\}^{K\times M^t}$, where an element $B_{k,m}=1$ indicates that AGV $k$ undertakes the transportation task $x_m$. The assignment determines the high-level task route for each AGV, denoted as $\mathcal{R}_k$. This task route is an ordered sequence of $R_k$ key physical locations that AGV must visit, formally expressed as $\mathcal{R}_k=[r_k(1),r_k(2),\cdots,r_k(R_k)]$. To formally link the global task index $m$ to its corresponding positions in the local route, we define a mapping. For any task $x_m$ assigned to AGV $k$, let $e_{m_+}$ and $e_{m_-}$ denote the indices of its pickup node $x_{m_+}$ and delivery node $x_{m_-}$ within the route $\mathcal{R}_k$, respectively. Thus $\mathcal{R}_k[e_{m_+}]=x_{m_+}$ and $\mathcal{R}_k[e_{m_-}]=x_{m_-}$. The task sequence $\mathcal{R}_k$ is constructed from the pickup and delivery nodes of both uncompleted tasks and new assigned tasks, along with resupply nodes.

\textbf{The definition of navigation route:} To execute the task sequence $\mathcal{R}_k$, the AGV then plans route scheduling and generates a detailed navigation route $\mathcal{P}_k$. This route is the ordered concatenation of the sub-paths that connect task nodes in the task route $\mathcal{R}_k$. Each sub-path, from a starting location $r_k(e)$ to a destination $r_k(e+1)$, is a sequence of $I_e+1$ adjacent vertices, $\mathcal{P}_k^e=[{p}_k^e(0),{p}_k^e(1),\dots,{p}_k^e(I_e)]$. Here, ${p}_k^e(0)=r_k(e)$, ${p}_k^e(I_e)=r_k(e+1)$, and $({p}_k^e(i),{p}_k^e(i+1))\in E$ for all $i=0,1,\dots,I_e-1$. 
Then, the entire navigation route of AGV $k$ is $\mathcal{P}_k=\mathcal{P}_k^0{\circ}\mathcal{P}_k^1{\circ}\cdots\circ{\mathcal{P}_k^{R_k-1}}$. At each time slot $t$, AGV takes a discrete navigation step $\widehat{a}_k^t$ from a vertex ${p}_k^e(i)$ to a linked adjacent vertex ${p}_k^e(i+1)$ or remains at the current vertex.

\textbf{Constraints and assumptions:} The common constraints and assumptions are summarized as follows.
1) The AGVs cannot pick up partially completed products until it is produced by the production robots \cite{2}; 2) Each AGV can transport only one task at a time \cite{17}; 3) Each task can only be transported by a single AGV \cite{tii}; 4) The loading and unloading time of the raw materials and partially completed products are included in the processing time of the task \cite{6}; 5) AGVs can load and unload material from any of the four sides of the production robot's grid with equal effect; 6) Task assignment for production MRS can be supported by the previous work \cite{2}, which is not the focus of this paper.

\subsection{Problem formulation of online scheduling}
Since a transportation task is considered complete when the AGV delivers the required quantity of materials for the production robot within the allotted time \cite{3}, we proceed to formulate the scheduling objective based on the completion times of all tasks. 
The completion time for a given task $x_m$ assigned to AGV $k$ is its arrival time at the task's delivery node, which is denoted by $\widehat{AT}_{k,e_{m^-}}$. The arrival time at any step $e$ in a route, $\widehat{AT}_{k,e}$, is calculated recursively, expressed as
\begin{equation}
\widehat{AT}_{k,e} = \widehat{AT}_{k,e-1} + \widehat{WT}_{k,e-1} + T(r_k(e-1), r_k(e)),\label{1}
\end{equation}
where $\widehat{WT}_{k,e-1}$ is the waiting time incurred at the previous node and $T(\cdot)$ is the travel time along the sub-path $\mathcal{P}_k^{e-1}$, which depends on the route decision vector $\widehat{\textbf{A}}_k$. Here, $\widehat{\textbf{A}}_k$ contains the sequence of navigation steps $\widehat{a}_k^t$ of AGV $k$ over time. The waiting time of AGV $k$ at a pickup node, indexed by $e_{m_+}$ for task $m$, is determined by the product's preparation time $\widehat{{PT}}_m$, denoted by $\widehat{WT}_{k,e_{m_+}} = \max\{0, \widehat{PT}_m - \widehat{AT}_{k, e_{m_+}}\}$. For all delivery and resupply nodes, the waiting time is defined as zero.

Crucially, dynamic interactions between production MRS and T-MRS exist due to the production-transportation task dependencies. AGV $k$ cannot pick up the partially completed product at node $x_{m_+}$ until two conditions are met: the necessary input materials that are the output of an upstream task have been delivered, and 2) the production robot at that node has finished its prior work \cite{4}. This dependency is captured by the preparation time of the partially completed product $\widehat{{PT}}_m$. 

To formalize this, for any task $x_m$ that is not the first in its production flow, we define its direct upstream task as $m_{up}$, whose priority is denoted by $\mu_{m_{up}}=\mu_m-1$. The AGV responsible for this upstream delivery is denoted as $k'$, and the completion time of this delivery is $\widehat{AT}_{k',e_{m_{up,-}}}$. Furthermore, let $m_{prev}$ be the task that was processed at the same production node as task $x_m$. The production robot becomes available only after the product for this prior task is ready. This time is given by the preparation time of the previous task, $\widehat{{PT}}_{m_{prev}}$. Thus, the preparation time  $\widehat{{PT}}_m$ is expressed as
\begin{equation}
\widehat{PT}_m = \max\{\widehat{PT}_{m_{prev}}, \widehat{AT}_{k', e_{m_{up,-}}}\} + t_{p,m},\label{2}
\end{equation}
where $t_{p,m}$ is the processing time of task $x_m$. Furthermore, to constrain the completion time of individual tasks, we define the tardiness for each task, denoted by $\widehat{TA}_{m} = \max\{0, \widehat{AT}_{k, e_{m_-}} - t_m^{ddl}\}$.

The makespan, defined as the latest completion time among all AGVs, effectively captures the overall scheduling performance in cyber-physical manufacturing systems \cite{makespan}, thus it is used as the objective to be minimized in this study. 
The makespan for completing all reconfigured tasks under partial observability is then naturally expressed as $\widehat{W} = \max \limits_{k\in \mathcal{K}, m\in \mathcal{X}^{{t}}} \{ B_{k,m} \cdot \widehat{AT}_{k,e_{m_-}}\}, t=1,2,\dots,T$.

Therefore, the optimization problem is formulated as
\begin{subequations}
     \begin{align}
            \min \limits_{\textbf{B},\widehat{\textbf{A}}} &\quad\quad \widehat{W} \label{4} \\
                \text{s.t. ~~}
                 &\sum_{k\in \mathcal{K}}{{B_{k,m}}}=1,  \forall m \in \{1, 2, \dots, M^t\},t\in T   \label{4a}\\
                & B_{k,m} \cdot \widehat{TA}_{m} \leq \tau_m,\forall k\in \mathcal{K},\forall m \in \{1, 2, \dots, M^t\},t\in T\label{4b}\\
                & B_{k,m} \cdot d_k^t \geq d_{m_-}, \forall k\in \mathcal{K},\forall m \in \{1, 2, \dots, M^t\},t\in T\label{4c}\\
                & B_{k,m} \cdot(d^*_k-d_k^t) \geq d_{m_+}, \forall k\in \mathcal{K},\forall m \in \{1, 2, \dots, M^t\},t\in T\label{4d}\\
                & B_{k,m} \in\{0,1\}, \forall k\in \mathcal{K},\forall m \in \{1, 2, \dots, M^t\},t\in T\label{4e}\\
                %& l_k^t \neq l_{k'}^t,\forall k\neq k'\in \mathcal{K},t\in T\label{4f}\\
                & K_{l_k^t}\leq \kappa^*, \forall k\neq k'\in \mathcal{K},t\in T\label{4g}
     \end{align}
    \end{subequations}
where $\widehat{\textbf{A}}=\{\widehat{\textbf{A}}_1,\widehat{\textbf{A}}_2,\dots,\widehat{\textbf{A}}_K\}$ represents the route decision matrix for all AGVs over time, where each route decision vector $\widehat{\textbf{A}}_k$ is derived from its local partial observation $\widehat{o}_k^t$ under partial observability. Constraint \eqref{4a} ensures that each task is conducted by only one AGV. Constraint \eqref{4b} guarantees that each task is completed within its due time, with a permissible soft delay $\tau_m$. Constraints \eqref{4c} and \eqref{4d} define the security payload and resupply conditions, respectively, for each AGV in practical manufacturing scenarios. 
Finally, constraint \eqref{4g} ensures congestion-free routes.

\section{Implementation for online scheduling of collision-free and congestion-free routes }

\subsection{Computationally feasible formulation of the problem}
\begin{figure}[ht]
	\centerline{\includegraphics[width=0.5\linewidth]{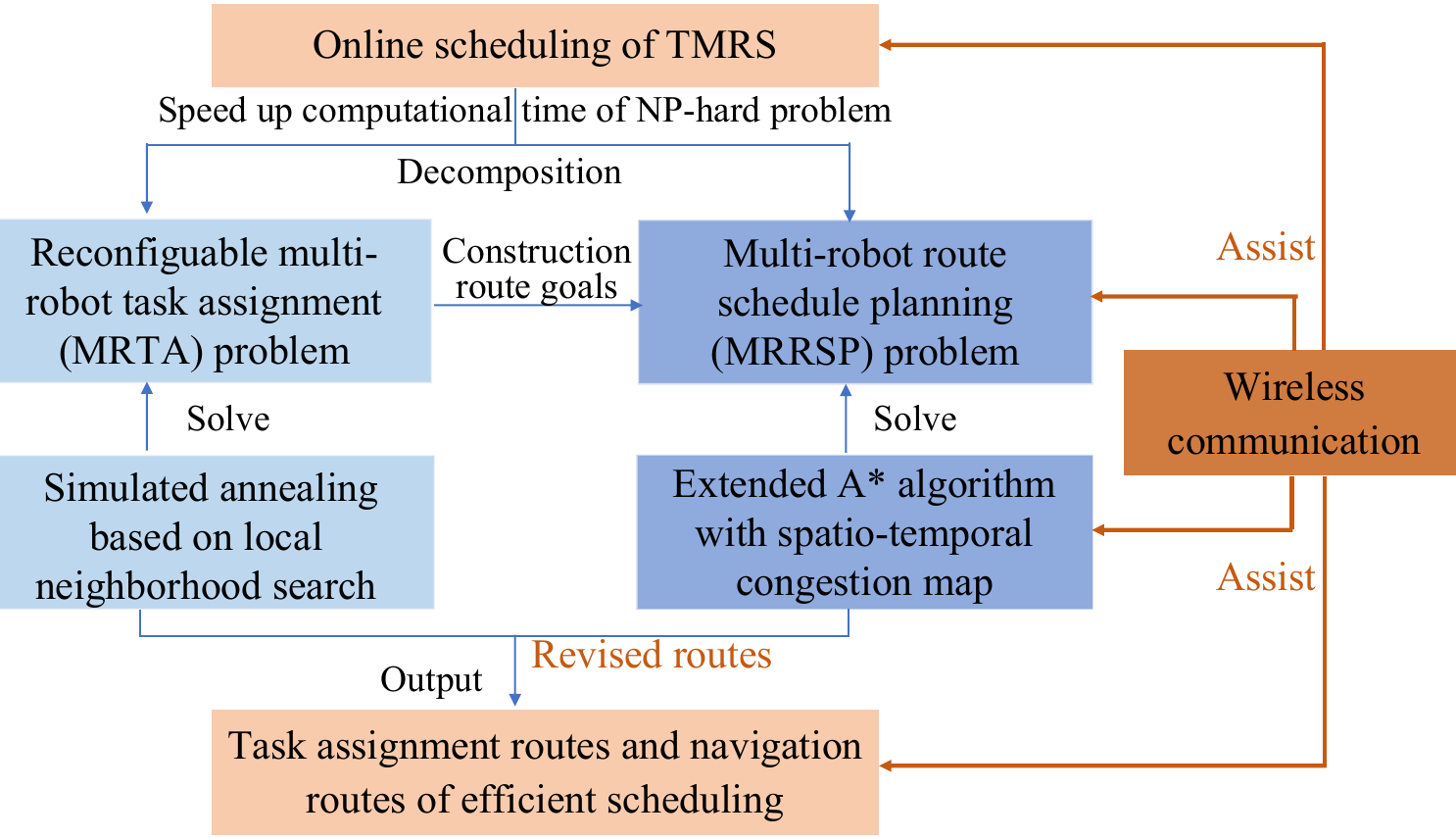}}
	\caption{The decomposition of the online scheduling problem and the coupling between the MRTA and MRRSP algorithms. 
    }
	\label{decom}
\end{figure}

The multi-robot scheduling problem is verified as NP-hard \cite{5,np2}, making exhaustive search infeasible for real-time operations. However, in a multi-robot factory where rapid decision-making is prioritized under dynamic environments, obtaining suboptimal solutions almost real-time is more effective and practical.
Therefore, to ensure the computational feasibility of online scheduling, we decompose the problem into two subproblems \cite{decom}: 1) a centralized MRTA problem to match transportation tasks to appropriate AGVs; and 2) a distributed multi-robot route scheduling planning (MRRSP) problem to schedule the route for each AGV. The logical framework of this decomposition and the interaction between the two subproblems are illustrated in Fig. \ref{decom}. The MRTA solution determines the routing objectives for MRRSP, while MRRSP generates paths that realize and assess the feasibility of MRTA decisions.
%The MRTA solution constructs the routing objectives for MRRSP, while MRRSP evaluates the decisions from MRTA.

However, the inherent partial observability of the global environment poses a significant challenge for multi-AGV coordination. Relying solely on its local observation $\widehat{o}_k^t$, an individual AGV lacks the situational awareness required to negotiate collision-free and congestion-free routes in a dynamic, multi-agent environment. To overcome this limitation, this paper leverages wireless communication to enable multiple AGVs to exchange intention information and thus compensate partial observability. 

\begin{figure}[ht]
	\centerline{\includegraphics[width=0.5\linewidth]{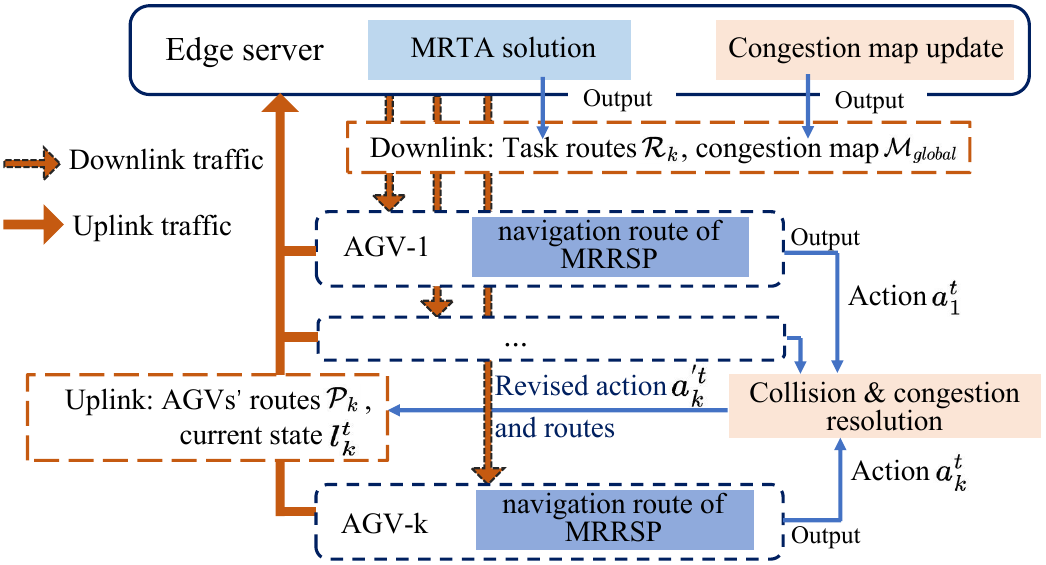}}
	\caption{The proposed communication-enabled online scheduling framework. 
    }
	\label{fig2}
\end{figure}

\subsection{Communication-enabled dynamic scheduling}
This section presents the dynamic scheduling scheme under the proposed communication-enabled online scheduling framework in Fig.~\ref{fig2}. The edge server centrally assigns MRTA solutions with a simulated annealing (SA) algorithm to AGVs and each AGV then executes MRRSP with the proposed modified A* algorithm in a distributed manner. During the execution process of AGVs, they exchange critical state and intention information via wireless communication, e.g., AGV's navigation path $\mathcal{P}_k^t$ and current state $l_k^t$. With such communicated information, denoted as $\tilde{o}_k^t$, each AGV forms an enhanced observation, which serves as a rich approximation of the global state $s^t$, enabling each AGV to dynamically revise its actions for safer and more efficient navigation.
The operation of two modules significantly relies on the specifically designed wireless network to exchange local observations to supplement partial observability and dynamically adjust routes. 
We elaborate on more details of the modular design in the following.

\subsubsection{ Simulated annealing for MRTA}
When dynamic production demands arrive, the edge server performs MRTA for both production MRS and T-MRS, respectively. Since MRTA for production MRS can be supported by the previous work in \cite{2}, this section focuses on MRTA for T-MRS, which determines the route goals for MRRSP. 
Due to the real-time computational demands of scheduling, heuristic search methods are practical for obtaining solutions within a short time \cite{np}. Thus, we utilize an SA algorithm with large neighborhood search (LNS) to solve MRTA. LNS, a widely used local search method \cite{17}, improves solution quality by iteratively destroying and repairing parts of a task assignment solution $\textbf{B}$. 

Specifically, each iteration consists of two parts that are guided by the tasks' priority, $\mu_m$: first, selecting a neighborhood to destroy by removing some tasks from the current solution $\textbf{B}$. The selection is not purely random but is biased by their position in the production flow. Tasks with a higher $\mu_m$ corresponding to a lower precedence, are assigned a higher probability of being removed.
Second, repairing it by reassigning the removed tasks to generate a new solution $\textbf{B}^{new}$. To ensure that precedence constraints are always satisfied, the removed tasks are first sorted by their sequence index $\mu_m$ and then inserted sequentially.
If the estimated completion time $\tilde{W}(\textbf{B}^{new})$ of the new solution is less than the current solution $\tilde{W}(\textbf{B})$, then $\textbf{B}^{new}$ is unconditionally accepted.
Otherwise, the probability of acceptance of $\textbf{B}^{new}$ is $p = \exp(\frac{-(\tilde{W}(\textbf{B}^{new}) - \tilde{W}(\textbf{B}))}{T^{{temp}}})$. SA uses a temperature $T^{{temp}}$ to determine the probability of accepting a non-improving solution, updating $T^{{temp}}$ with $\alpha \times T^{{temp}}$ after each iteration. In this scheme, consistent with \cite{17}, we use the term ``estimated'' time to indicate the time calculated under the assumption that all AGVs follow their shortest paths on the factory that ignore the congestions and collisions between each other. As analyzed earlier, for a given task assignment solution $\textbf{B}$, each AGV $k$ is assigned a task route $\mathcal{R}_k=[r_k(1),r_k(2),...,r_k(R_k)]$. The estimated completion time for this AGV, $\tilde{t}_k$ is calculated as the sum of shortest path travel times, starting from its initial position and executing all tasks in sequence. The estimated completion time $\tilde{W}(\textbf{B})$ of the solution is defined as the maximum estimated completion time among all AGVs: $\tilde{W}(\textbf{B})=\max_{k\in\mathcal{K}}\{\tilde{t}_k\}$. 
After the central edge server conducts the search algorithm for MRTA, it transmits the corresponding task route $\mathcal{R}_k$ to each AGV $k$ via downlink wireless communication. 
\begin{algorithm}[t]
    \renewcommand{\algorithmicrequire}{\textbf{Input:}}
    \renewcommand{\algorithmicensure}{\textbf{Output:}}
    \caption{MRTA strategy based on LNS-assisted SA}
    \label{alg:1}
    \begin{algorithmic}
        \REQUIRE Initial task assignment solution $\textbf{B}$, initial temperature $T^{{temp}}$, termination temperature $T^{{temp*}}$, and cooling factor $\alpha$. Set the objective function \eqref{4} to $\tilde{W}(\textbf{B})$. 
        \ENSURE Best task routes solution $\hat{\textbf{B}}$;
        \STATE Initialize best solution $\hat{\textbf{B}}\leftarrow \textbf{B}$;
        \WHILE{$T^{{temp}} >T^{{temp*}}$}
    \STATE Choose a neighborhood to destroy by removing some tasks from $\textbf{B}$;
     \STATE Repair a neighborhood by reassigning the removed tasks to generate a new solution $\textbf{B}^{new}$;
\STATE Obtain the estimated completion time ;
            \IF {$\tilde{W}(\textbf{B}^{new}) < \tilde{W}(\textbf{B})$}
                \STATE Set $\textbf{B} \leftarrow \textbf{B}^{new}$;
            \ELSE
                \STATE Set $\textbf{B} \leftarrow \textbf{B}^{new}$ with probability $p$;
                \STATE $p = \exp\left(\frac{-(\tilde{W}(\textbf{B}^{new}) - \tilde{W}(\textbf{B}))}{T^{{temp}}}\right)$;
            \ENDIF
            \IF {$\tilde{W}(\textbf{B}) < \tilde{W}(\hat{\textbf{B}})$}
                \STATE Set $\hat{\textbf{B}} \leftarrow \textbf{B}$, 
            \ENDIF
            
            \STATE Reduce temperate $T^{{temp}} \leftarrow \alpha \times T^{{temp}}$;  
        \ENDWHILE
        \RETURN $\hat{\textbf{B}}$
    \end{algorithmic}
\end{algorithm}

\subsubsection{Congestion map-aware A* for MRRSP}
Based on the received task routes via wireless networking, each AGV executes distributed MRRSP to transport industrial materials for production robots. However, when multiple AGVs traverse the intersection segment simultaneously in different directions, congestion can lead to delays \cite{congestion-aware1}. 
While the traditional A* algorithm is effective at finding the shortest spatial path \cite{19}, it often neglects the temporal dynamics of multi-AGV traffic, leading to congestion. Our proposed method addresses this by extending A* to a spatiotemporal domain, guided by a spatiotemporal congestion map, denoted as $\mathcal{M}_{global}$.

This map is generated and periodically updated by a central server, which predicts the number of AGVs occupying each node for future time slots, enabling AGVs to obtain an approximate, system‑wide observation. Upon receiving this congestion map via the wireless downlink, each AGV dynamically adjusts its route to proactively avoid potential congestion by taking detours.

To incorporate time-varying congestion, the extended A* algorithm searches over a node state space where each state $n$ is a tuple $(v,t)$,  representing the arrival at a physical node $v$ at time slot $t$. When an AGV $k$ needs to plan a path from its current location $l_k^t$, the A* search is initiated with the start state $n_\text{start}=(l_k^t,t)$. The algorithm then explores successor states $n'=(v',t')$ to find an optimal path by minimizing the total cost function $\phi(n')$: $\phi(n')=g(n')+h(n')+c(n')$. Here, $g(n')$ is the actual cost, representing the travel time from the start state $n_\text{start}$ to the current state $n'$. $h(n')$ is heuristic function that estimates the minimum remaining travel time derived from Manhattan distance. $c(n')$ is the accumulated congestion penalties, which is determined from the congestion map: $\theta(v', t') = \begin{cases} P_{\text{congestion}}, & \text{if } \mathcal{M}_{global}(v', t') \geq \kappa^* \\ 0, & \text{otherwise} \end{cases}$. Here, $P_{\text{congestion}}$ is a congestion cost value, which ensures that the generated navigation route proactively bypasses high-density areas, thereby achieving a congestion-free path at the planning level. $ \kappa^* $ is the node congestion threshold, which determines the maximum allowable AGV density at a node to prevent deadlocks. In this study, we set $\kappa^*=3$ based on the empirical analysis in \cite{3}, which indicates that the collision probability exceeds 90\% when three or more AGVs occupy the same location, posing a severe risk to navigation safety.
 By minimizing this composite cost function, the modified A* algorithm balances path length and congestion cost, resulting in more efficient navigation.
Upon reaching the goal $r_k(e)$, the reach time $\widehat{AT}(r_k(e))$ is computed and updated according to \eqref{1}. If $\widehat{AT}(r_k(e))-t^{ddl}_{r_k(e)}\leq \tau_e$, the path is reconstructed by tracing back through the parent nodes. By introducing dynamic congestion map, this method enables AGVs to avoid congestion dynamically rather than following the shortest path, as shown in Algorithm~2.

\begin{algorithm}[h]
    \renewcommand{\algorithmicrequire}{\textbf{Input:}}
    \renewcommand{\algorithmicensure}{\textbf{Output:}}
    \caption{MRRSP design based on congestion map-aware A*}
    \label{alg:a_star}
    \begin{algorithmic} %Task assignment solution $\textbf{B}$ constructed by MRTA,
        \REQUIRE Start location $l_k^t$, goal location $r_k(e)$, current time $t$, congestion map $\mathcal{M}_{global}$, due time of goal node $t^{ddl}_{r_k(e)}$, and soft delay of goal node $\tau_e$
        \ENSURE Path from $l_k^t$ to $r_k(e)$
        \STATE Initialize open\_list, closed\_list
        \STATE Push start node $n_{start}=(l_k^t,t)$ into open\_list with $g(n) \leftarrow 0$, $\widehat{AT}(n) \leftarrow t$
        \WHILE{open\_list not empty}
         \STATE Pop $n=(v,t)$ from open\_list
        \IF{$v == r_k(e)$ }
         \IF{ $r_k(e)$ is a pickup node}
         \STATE \textbf{Compute:} $PT(r_k(e))$ by \eqref{2} and $\widehat{WT}(r_k(e)) \leftarrow \max\{0, \widehat{PT}(r_k(e)) - \widehat{AT}(r_k(e))\}$
                \ELSE
                \STATE Set waiting time $\widehat{WT}(r_k(e)) \leftarrow 0$
                \ENDIF
                 \STATE \textbf{Compute:} $\widehat{AT}(r_k(e))$ according to \eqref{1}
               
                \IF {$\widehat{AT}(r_k(e))-t^{ddl}_{r_k(e)}\leq \tau_e$}
                  \RETURN Reconstruct path
                  \ELSE
                   \STATE Continue
                \ENDIF
             \ENDIF
            
            \STATE Add $n$ to closed\_list
            \FOR{each neighbor $v'$ of $v$}
               \STATE $t'\leftarrow t+1$, $n'\leftarrow(v',t')$
                \IF{$n'$ is invalid or in closed\_list}
                    \STATE Continue
                \ENDIF
                \STATE Update $h(n') \leftarrow \text{Manhattan\_Distance}(v', r_k(e))$, $g(n') \leftarrow g(n) + 1$, and $c(n') \leftarrow c(n)+\theta(n')$
                
                \STATE \textbf{Calculate total cost for $n'$:} $\phi(n') \leftarrow g(n') + h(n') + c(n')$
                
                \IF{neighbor node is not in open\_list or $\phi(n')$ is smaller than the existing cost}
                \STATE Push neighbor node $n'$ to open\_list
                 
                \STATE Update cost and parent of $n'$ to $n$ in open\_list
            \ENDIF
            \ENDFOR
        \ENDWHILE
        \RETURN No path found
    \end{algorithmic}
\end{algorithm}

\section{Wireless networking suggested by online scheduling}
This section focuses on wireless networking design to meet the requirements of online scheduling.
Wireless networking will be developed on two aspects: \textit{a)} determining the required information to supplement partial observability, and \textit{b)} designing the wireless networking framework, radio resources, and transmission process tailored to scheduling requirements. Finally, the uplink success probability and throughput for a finite number of AGVs under the proposed framework are derived, providing a theoretical basis for evaluating the overall online scheduling scheme.

\subsection{Effective communication traffic for online scheduling of T-MRS}
We define the following as the critical M2M traffic for our online scheduling framework.
\begin{itemize}

\item[$\bullet$] \textbf{Current state $l_k$}: An uplink packet transmitted by AGV $k$ every $D_l$ time slots with size $S_{l}$ bytes, which conforms to a constant bit rate (CBR) pattern. It reflects the two-dimensional location of AGV at current time, which supports real-time scheduling adjustments.
\item[$\bullet$] \textbf{Navigation route $\mathcal{P}_k$}: An uplink packet sent by AGV $k$ every $D_{\mathcal{P}}$ time slots, with size $S_{\mathcal{P}}$ bytes, and forwarded to other AGVs via APs. This traffic conforms to a CBR pattern. It encodes the planned two-dimensional trajectory as a time-indexed sequence, which reflects the intention information of AGV.
\item[$\bullet$] \textbf{Global spatiotemporal congestion map $\mathcal{M}_{global}$}: A downlink packet is generated and sent by the edge server with an interval of $D_{\mathcal{M}}$ time slots with size $S_{\mathcal{M}}$ bytes, conforming to a CBR pattern and containing predictive congestion nodes across the factory with timestamps. Constructed using received navigation routes, the congestion information guides AGVs along congestion-free paths. 
\item \textbf{Task routes $\mathcal{R}_{k}$}: A downlink packet of size $S_{\mathcal{R}}$ bytes is sent when the edge server completes MRTA, producing variable bit rate downlink traffic due to randomly arriving tasks.
\end{itemize}

\subsection{Wireless networking framework, radio resources, and transmission process }
As depicted in Fig. \ref{fig1}, the network framework comprises three primary components: 1)  The edge server is responsible for centralized scheduling and management. It controls all APs, providing global fused information based on the routes and local states of AGVs; 2) Each AP provides wireless connectivity to the AGVs within its communication range, facilitating the transmission of uplink and downlink traffic; 3) High-bandwidth, low-latency optical fiber links connect the APs to the edge server, ensuring that data is rapidly forwarded from the APs to the central server, forming a reliable backbone for the entire network. 

Aligned with the 3GPP physical layer specification \cite{3gpp}, a resource element (RE) is defined as the transmission of one symbol over one subcarrier, representing the smallest unit on the time-frequency grid. Suppose that orthogonal frequency division multiple access (OFDMA) is used for the physical layer transmission.
A set of frequency-adjacent REs is aggregated into a resource block (RB), which serves as the basic unit for resource allocation. Each time slot contains a fixed number of REs. We define a group of REs with a consistent shape and position on the time-frequency grid as a radio resource unit (RRU), and each RRU is mapped to a logical channel. Throughout the paper, the term ``channel" refers to this ``logical channel" for brevity. 
For simplicity, it is assumed that each RRU/ channel can accommodate one data packet per time slot for transmission, e.g., a route packet with $S_{\mathcal{P}}$ bytes and a state packet with $S_{l}$ bytes. Additionally, the available radio spectrum is assumed to support $C$ orthogonal channels.

\textbf{Uplink real-time contention-based multi-link transmission:} Since AGVs must adjust their navigation actions in real time based on communicated observations to avoid collisions and congestion, low-latency communication is essential for timely and effective decision-making in a smart factory \cite{8}. Retransmissions are often ineffective in highly dynamic environments \cite{lyx}. Thus, a retransmission-free contention-based access protocol is employed to support online scheduling, which is akin to real-time ALOHA by grant-free access, no acknowledgment, and discarding retransmission \cite{14}.
Furthermore, to enhance transmission reliability, multi-link transmission is utilized, where the same packet is simultaneously transmitted on multiple RRUs/channels. If multiple AGVs select the same channels simultaneously, cyber collisions are inevitable. Therefore, a packet is successfully transmitted when at least one copy is sent without collision. 
It should be noted that our system design is inspired by the multi-link operation specified in Wi-Fi 7 \cite{802.11,MLO}, which enables simultaneous data transmission across multiple frequency bands. Without loss of generality, we assume that error control is available so that we can focus on collisions caused by simultaneous packet transmissions over the same link.

Therefore, we design the contention-based multi-link transmission mechanism without retransmission, which is tailored for AGV scheduling traffic, where timeliness outweighs absolute reliability. 
By leveraging multi-link redundancy to exploit frequency diversity, it effectively mitigates the vulnerability of single-link transmission against cyber collisions and errors. This makes it both effective and well-suited for MRS operations in smart factories.
For an uplink data packet, AGV randomly selects $S$ channels to transmit the same packet simultaneously to the AP, e.g., navigation route $\mathcal{P}_k$ and current state $l_k^t$. The AP receives and decodes the packet based on the selection combining strategy while abandoning all retransmission and acknowledgment processes. The successfully decoded data is then forwarded to the edge server for centralized processing.

\textbf{Downlink transmission process:} During the downlink transmission process, the edge server transmits a global spatiotemporal congestion map  $\mathcal{M}_{global}$ to AP. Then, AP broadcasts the current congestion map to AGVs.

\subsection{Uplink transmission performance analysis}

This subsection analyzes the uplink success probability for the contention-based retransmission-free multi-link transmission scheme from a networking perspective, which helps understand how the real-time multi-link transmission networking affects AGV scheduling. Note that when $S$ RRUs or channels are selected to transmit a packet, a collision occurs if all these RRUs collide with other transmissions.
For simplicity, each AP is assumed to provide full communication coverage across the entire factory area, thus eliminating the need to consider the handover process. Although $Q$ APs are uniformly deployed, they are assumed to logically function as a single AP. All AGVs compete for a common set of $C$ orthogonal channels. Consequently, the system behaves identically to a single-AP setup without spatial frequency reuse.

Assume each AGV transmits one uplink packet on average every $D$ time slots.
To model this probabilistically, we treat the transmission behavior as a Bernoulli process, where each AGV independently attempts to transmit in each time slot with probability $P_t$. Thus, the expected number of transmissions over $D$ slots is $\mathbb{E}(X) = D \cdot P_t = 1$, where $X$ is the number of transmissions.

We initially consider two AGVs, denoted by $a$ and $b$. Let $i$ be the number of colliding channels between them. AGV $a$ can successfully transmit under two mutually exclusive conditions: 1) AGV $b$ is idle (i.e., not transmitting), which occurs with probability $1-P_t$. 2) AGV $b$ attempts to transmit, but its transmission collides with only $i$ of the $S$ channels selected by AGV $a$, which occurs with probability $P_t\frac{\binom{C-i}{S}}{\binom{C}{S}}$. Thus, the probability of successful transmission without collisions is the sum of these two probabilities: $ P_t \cdot {\left ((1-P_t)+P_t \frac{\binom{C-i}{S}}{\binom{C}{S}}\right )}$. 
Based on this, it can be extended to more than two AGVs. Since there are $K-1$ independent AGVs, the probability extends to: $P_t \cdot {\left ( (1-P_t)+P_t \frac{\binom{C-i}{S}}{\binom{C}{S}}\right )}^{K-1}$.

When considering all subsets of channels, the \textit{Inclusion-Exclusion Principle} is used to prevent over-counting of collision occurrences \cite{IEP}. Therefore, summing over $i$ from 1 to S yields:
\begin{equation}
 P_{success}=\sum_{i=1}^{S}(-1)^{i+1}\binom{S}{i}\cdot P_t \cdot{\left [(1-P_t)+P_t\frac{\binom{C-i}{S}}{\binom{C}{S}}\right ]}^{K-1}. \label{analyze}
\end{equation}

Then, the throughput of the transmission process is given by:
\begin{equation}
\eta=\sum_{i=1}^{S}(-1)^{i+1}\binom{S}{i}\cdot K\cdot P_t \cdot{\left [(1-P_t)+P_t\frac{\binom{C-i}{S}}{\binom{C}{S}}\right ]}^{K-1}. \label{analyze1}
\end{equation}

\begin{figure*}[!t]
\centering

\subfloat[]{
		\includegraphics[scale=0.3]{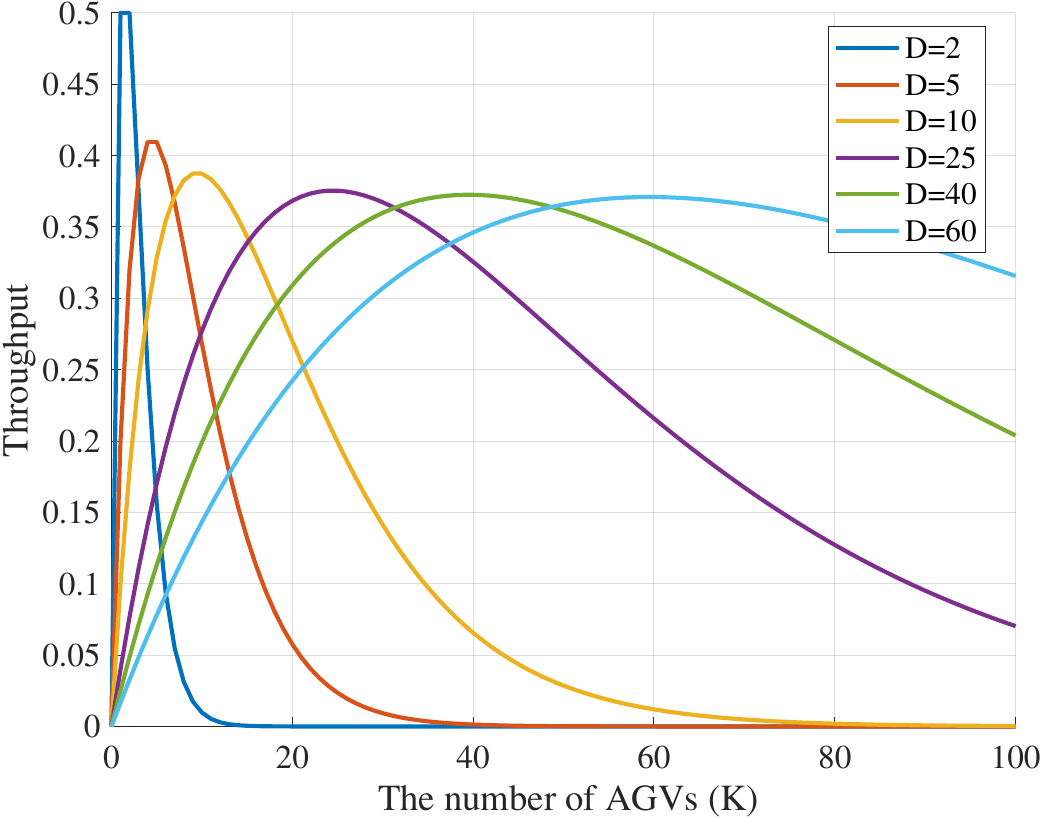}}      
\subfloat[]{
		\includegraphics[scale=0.3]{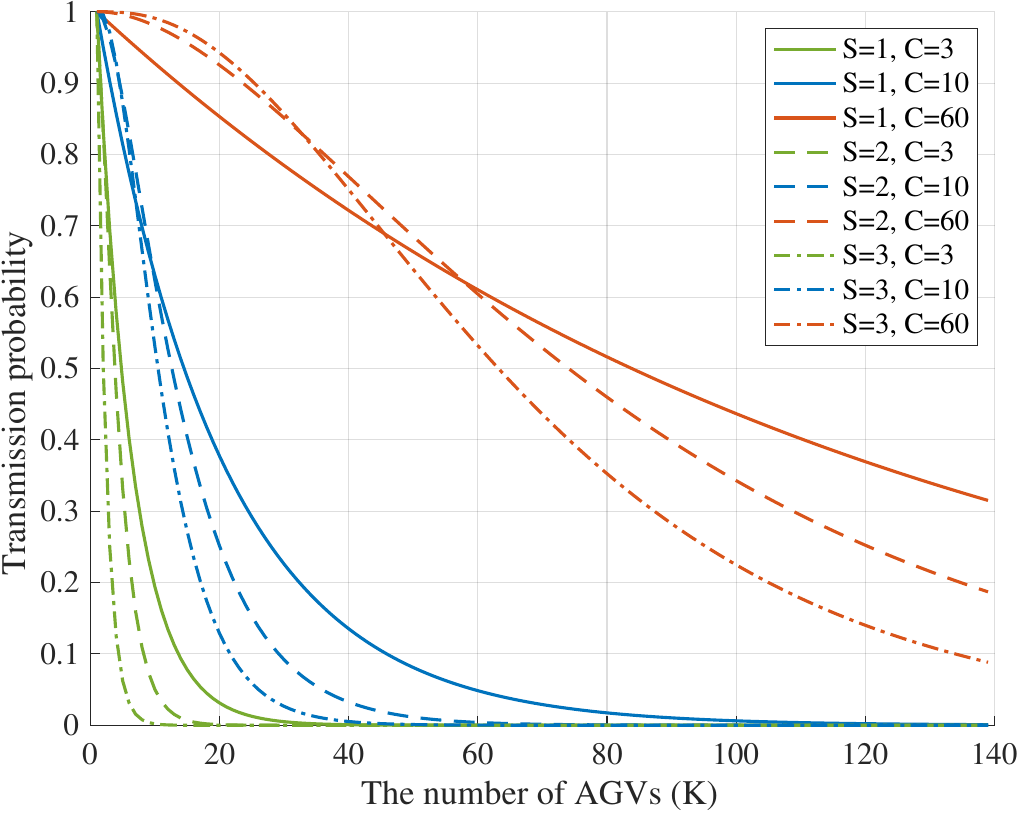}}
\subfloat[]{
		\includegraphics[scale=0.3]{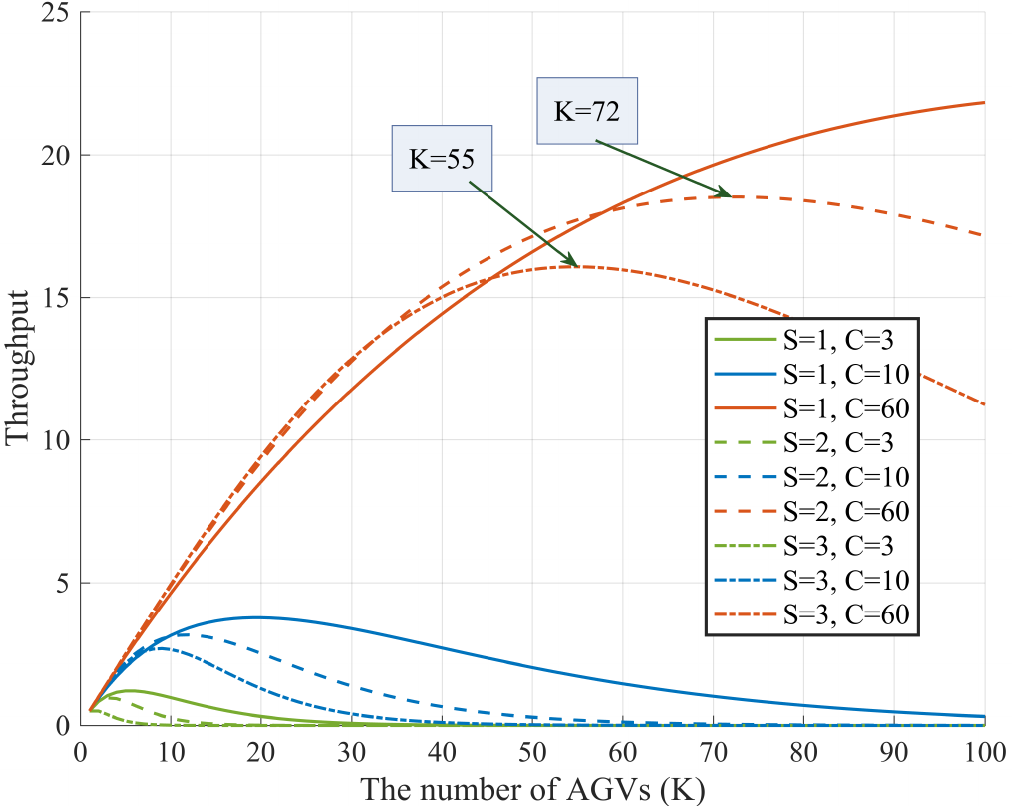}}
\caption{\raggedright Comparison of the performances of transmission probability and throughput under different numbers of AGVs. (a) Throughput vs. the number of AGVs for different D values when $C=S=1$; (b) Transmission success probability vs. the number of AGVs for different selected number of channels S when $D=2$; (c) Throughput vs. the number of AGVs for different selected number of channels S when $D=2$.}
\label{fig4}
\end{figure*}

Before moving to the performance evaluation of the designed overall online scheduling scheme, we proceed to analyze the derived expressions of formulas (\ref{analyze}) and (\ref{analyze1}) for the probability of successful transmission and the corresponding throughput. The results in Fig. \ref{fig4}(a) indicate that for a system with high AGV density and limited communication resources ($C=1,S=1$), throughput can be improved by enlarging the communication interval, as this alleviates cyber congestion. The results in Figs. \ref{fig4}(b) and \ref{fig4}(c) further demonstrate that when the AGV-to-channel ratio is low, multi-link transmission significantly enhances reliability by offering redundancy across multiple channels, thereby mitigating single-link interference. 
However, as the number of AGVs increases, see Fig. \ref{fig4}(c), the advantages are gradually offset by increased channel contention. This follows from the binomial factor ${\binom{C-i}{S}}/{\binom{C}{S}}$ in (\ref{analyze1}), which causes faster decay of transmission success probability with larger $S$. As a result, although multi-link is advantageous at moderate AGV-to-channel ratios, single-link eventually outperforms it. For $C=60$ and $D=2$, this crossover is observed at $K=72$, as shown in Fig. \ref{fig4}(c). At an ultra-high number of AGVs, the system breaks down where both transmission success probability and throughput approach zero. 
The results indicate that while multi-link transmission is beneficial, under high AGV-to-channel ratios, the system should adopt fewer channels $S$ or a larger communication interval $D$ to enhance transmission reliability. These results provide a baseline for comparison with the results of scheduling-oriented communication for T-MRS.

\renewcommand{\thesubfigure}{(\alph{subfigure})} 

\subsection{ Integration of  wireless networking and dynamic scheduling} 
Based on the designed wireless networking, this subsection illustrates how communicated information mitigates the limitation of partial observability within the proposed online scheduling mechanism. The mechanism forms a closed loop of uplink reporting, centralized processing, downlink broadcast, distributed path execution and dynamic replanning.

\textbf{Uplink phase:} Each AGV $k$ transmits its current state $l_k^t$ and navigation route $\mathcal{P}_k^t$ to the associated AP via contention-based multi-link transmission at every $D$ time slots. The AP then forwards the successfully received reports to the edge server.

\textbf{Edge server processing and downlink phase:} Upon the arrival of new task demands, the edge server first executes the MRTA procedure using the Algorithm 1 to assign tasks and generate task routes $\mathcal{R}_k^t$ for each AGV. Concurrently, the edge server constructs the global spatiotemporal congestion map $\mathcal{M}_{global}^t$ from uplinked information. The updated congestion map and navigation route are then transmitted via downlink to the AGVs that successfully reported their information, enabling them to incorporate global networked information into their route planning. 

\textbf{Distributed AGVs decision:} Each AGV operates in a distributed path execution and dynamic replanning process. Each AGV observes nearby AGVs within its sensing range $r$ and forms a local map $\mathcal{M}_{local,k}^{t}$. Upon receiving global map, the AGV merges $\mathcal{M}_{global}^{t}$ with $\mathcal{M}_{local,k}^{t}$ via an element-wise maximum operator.
Subsequently, the AGV executes the MRRSP using the congestion map-aware A* algorithm (Algorithm 2). Specifically, strictly following the mechanism defined in Section 3.2.2, the algorithm incorporates the merged congestion information into its cost function. By assigning high penalty costs to nodes where the predicted occupancy exceeds the congestion threshold $\kappa^*$, the AGV proactively generates a navigation route that bypasses these high-traffic areas.

With the generated route, AGV determines the planned action $a_k^t$. Before executing the action, the AGV performs a real-time conflict detection by examining the received routes for overlapping future positions. Conflicts are then resolved through predefined right-of-way rules: an AGV with a lower priority needs to stay and wait until all higher-priority AGVs have cleared the conflict area. {\small\textcircled{\footnotesize 1}} highest priority: the northbound AGV; 
{\small\textcircled{\footnotesize 2}} second priority: the southbound AGV;
{\small\textcircled{\footnotesize 3}} third priority: the eastbound AGV;
{\small\textcircled{\footnotesize 4}} lowest priority: the westbound AGV. 
In the event of communication failure due to severe packet collisions or errors, the AGV reverts to a local reasoning mode. Specifically, if no new congestion map is available, each AGV only relies on local sensing within its sensing range $r$ and forms a local map $\mathcal{M}_{local,k}^{t}$. The AGV then merges this with its most recently received global map $\mathcal{M}_{global}^{t^{\prime}}$, which retains the historical reservations of other AGVs, to perform MRRSP. Critically, the physical safety is guaranteed by the right-of-way rules: if another AGV is detected and the current AGV does not have right-of-way, it stays until the higher-priority AGV passes for collision resolution.

In summary, each AGV first follows the planned full-path generated by congestion-aware A*. The priority-based right-of-way rules are triggered only when an imminent collision risk is detected.
Only when these rules lead to a deviation, i.e., the actual action no longer follows the originally planned path, is Algorithm 2 invoked again to replan a new route. 
This design enables AGVs to adaptively adjust their navigation with real-time wireless communication, allowing proactive congestion avoidance and collision mitigation, which reduces unnecessary waiting and improves overall scheduling efficiency compared to purely local, perception-based navigation.

\subsection{Complexity analysis of distributed route scheduling}

As MRTA is centrally managed by the edge server, this subsection analyzes the computational complexity of distributed MRRSP, which is executed by each AGV under local resource constraints. In the communication-enabled scheduling scheme, the major local computational tasks of each AGV include 1) route scheduling based on the modified A* method with the received global congestion map, with computational complexity $O(Vlog(V))$, and 2) collision detection and resolution based on the global congestion map against neighboring AGVs $\mathcal{N}_k^t(r)$, with complexity $O(|\mathcal{N}_k^t(r)|)$. Therefore, the computational complexity of each AGV in the proposed communication-enabled scheduling scheme is $O(Vlog(V) + |\mathcal{N}_k^t(r)|)$.

In the traditional local reasoning-based scheme, the main local computational tasks of each AGV include 1) the generation of local congestion maps with computational complexity $O(|V_k(r)|)$, 2) independent route planning with a modified A* method, whose computational complexity is $O(Vlog(V))$, and 3) pairwise collision detection and resolution with other AGVs due to the lack of shared states, whose computational complexity is $O(|\mathcal{N}_k^t(r)|^2)$. Therefore, the computational complexity of the local reasoning-based scheme is $O(|V_k(r)| +Vlog(V) + |\mathcal{N}_k^t(r)|^2)$.
The proposed scheme leverages wireless networking to share the global congestion map generated by the edge server. This eliminates the need for each AGV to compute and predict other AGVs' intentions or routes in real-time and effectively reduces the local computational burden. 
Quantitatively, for a factory grid (e.g., $50 \times 50$), the dominant computation $O(V \log V)$ executes in a few milliseconds on standard embedded processors (e.g., 100 MFLOPS). This negligible latency, compared to the typical 100–500 ms control cycle, guarantees online feasibility.

\section{Computational experiments}
This section first demonstrates the necessity of wireless networking in enabling online scheduling under partial observability, then evaluates the impact of key communication resource settings on the scheduling efficiency of T-MRS. Finally, it discusses the insights gained from the simulations. We implement our approach with Python 3.8 on a server configured with Intel\textregistered\ Xeon\textregistered\ Platinum 8369B CPU.
The experiments are conducted in a $10\times10$ grid with AGV numbers $K$ ranging from 2 to 60 for typical scenarios, and extended up to 100 to evaluate the system performance under extreme loads.
The task set includes 60 production lines, each consisting of 4 tasks, which delineate the locations of the AGV’s pickup and delivery nodes. 
The raw material demands and the processing time for each task are generated randomly following a uniform distribution $U(5,10)$. This setting is chosen to verify the general applicability of the proposed scheme across diverse load conditions without assuming specific production patterns or avoiding introducing any specific bias.
 Each experiment is repeated five times with distinct seed sets for the environment and each AGV, and the results are averaged across these runs to ensure statistical reliability. The threshold $ \kappa^*$ for free congestion is set as 3, which aligns with \cite{3}. The maximum payload $d^*$, the perception range $r$ of AGVs, and the cooling factor $\alpha$ are set as 20, 2, and 0.995, respectively. In the communication configuration, different types of data packets are transmitted with the same communication interval. While most prior communication-centric studies focus on optimizing wireless performance alone, our work integrates communication and scheduling, treating communication as a core driver to enhance scheduling efficiency. This fundamental difference in objective makes a direct comparison with methods optimized for traditional wireless performance metrics non-trivial. This paper aims to demonstrate the the impacts from wireless communication on the online scheduling of T-MRS.  

\subsection{Ideal wireless networking supports online scheduling}
\begin{figure}[t]
	\centering
    \includegraphics[width=0.4\linewidth]{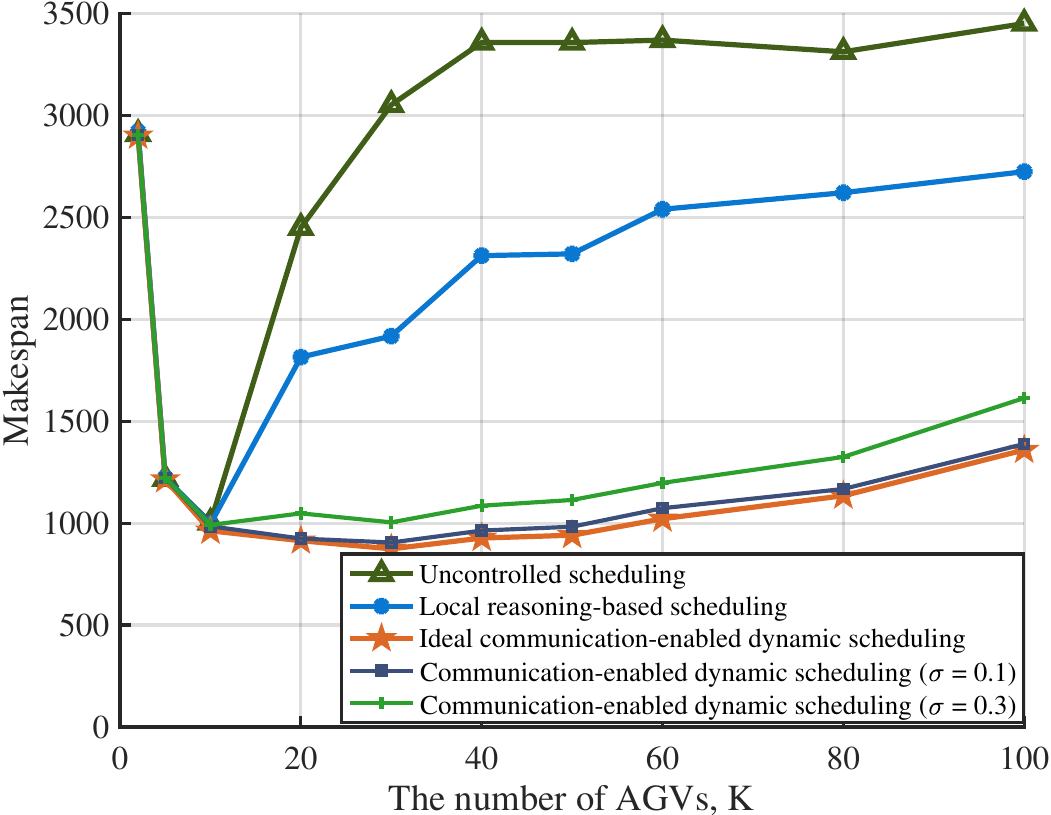}
\caption{\raggedright The number of AGVs vs. makespan}
\label{fig5}
\end{figure}

We first show the superiority of the proposed scheme for online scheduling under ideal communication conditions (infinite resources, no cyber collision, and no errors), see Fig.~\ref{fig5}. To emulate dynamic production, task sets are reconfigured three times without prior knowledge to the AGVs. Three baselines are considered: two non-communication and one error-prone communication scheme: 1) An uncontrolled scheduling scheme that is used to expose the occurrence of collisions and congestion in a smart factory. 2) A local reasoning-based scheme that relies solely on local partial observations and requires two additional time slots to stay for navigation safety. 3) A communication-enabled scheme without retransmissions, where random errors occur independently at each uplink link with probability $\sigma$.

As depicted in Fig. \ref{fig5}, increasing the number of AGVs initially improves the production process by enabling more parallel production. However, beyond 10 AGVs, significant performance degradation occurs under the uncontrolled baseline. The whole trend is primarily due to physical collisions and congestion that introduce substantial delays and disrupt factory operations. Similarly, with the local reasoning-based scheme, the makespan grows rapidly beyond 10 AGVs, indicating that large-scale AGV deployment is unsustainable when relying only on local observations. 
In contrast, the proposed communication-enabled scheduling substantially improves scalability. Even when the number of AGVs exceeds 30, the makespan remains significantly lower, achieving a 54.42\% efficiency gain compared to the local reasoning baseline. This gain is attributed to real-time updates of the global congestion map $\mathcal{M}_{global}$ and route exchange via wireless networking, which alleviate the effects of partial observability and enable AGVs to proactively adjust their navigation to avoid predicted congestion. It should be noted that the observed turning point cannot be determined by a closed-form expression. Rather, it results from the coupled effects of wireless information exchange, AGV traffic density, task reconfiguration, and factory size. While the exact threshold is scenario-dependent, the fundamental trend is robust: system efficiency improves with moderate AGV density due to parallel production, but beyond a certain level, physical congestion leads to performance degradation. 

Furthermore, Fig.~\ref{fig5} illustrates the impact of communication errors. When random errors occur with $\sigma=0.1$ and $\sigma=0.3$, performance degradation is observed compared to the ideal communication case. However, the reduction is notably smaller than in the non-communication baselines. The rationale is that our scheduling scheme adopts the multi-link duplication, effectively mitigating single-link error and enhancing transmission reliability.

These results consistently indicate that real-time wireless networking is indispensable for online scheduling under partial observability. Without communication, AGVs relying on local partial observations incur significant delays due to collisions and congestion, making factory operations prone to severe disruptions.

\subsection{Online scheduling with realistic wireless networking}
In a practical smart factory with limited radio resources, wireless networking is subject to cyber contention, leading to information loss. Since each AGV relies on received observations for proactive decision-making, such information loss reintroduces partial observability. Thus, we investigate how different wireless network configurations affect the overall factory performance. 

\textbf{The effects of communication interval $D$:} We first evaluate how the communication interval $D$ impacts overall scheduling performance. A resource-constrained scenario is simulated with 60 AGVs sharing a single channel ($C=S=1$), where cyber contention is most severe. This setting aims to further show the effectiveness and necessity of wireless networking in supporting scheduling performance even with limited resource.
 %9.53\% 

\begin{figure}[t]

\centering
\includegraphics[width=0.38\linewidth]{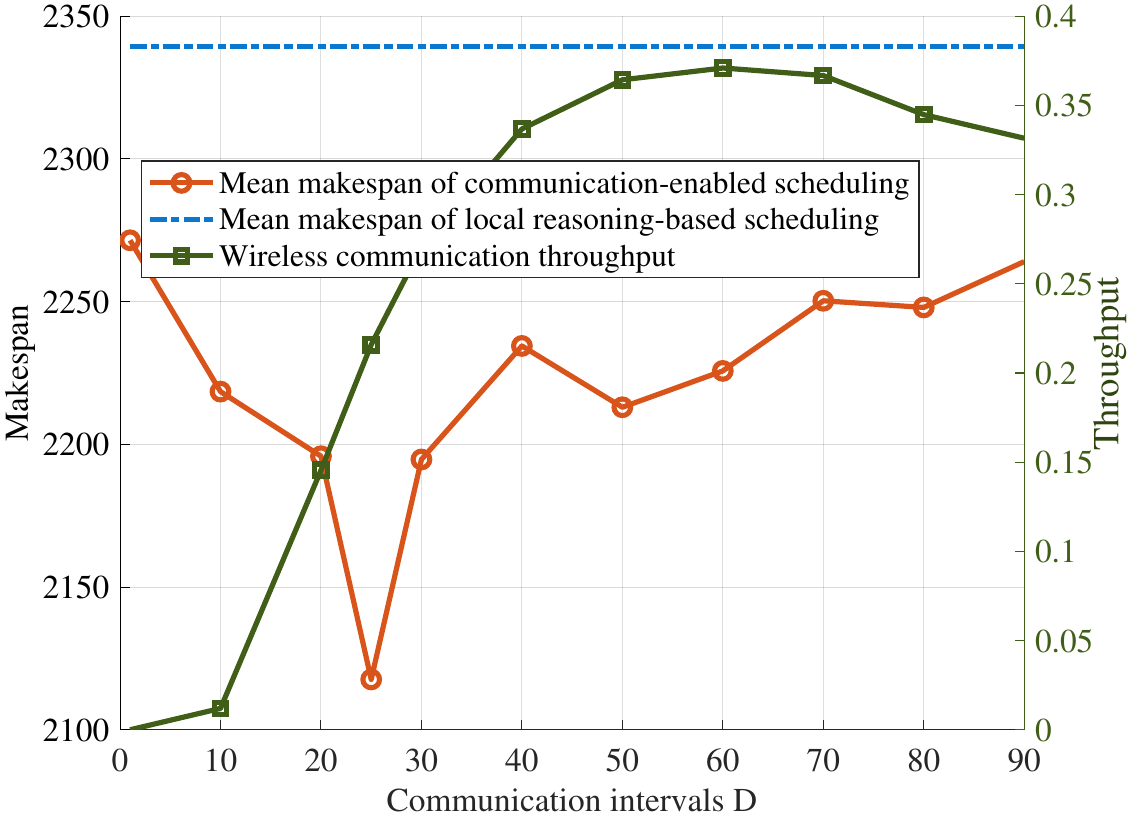}

\caption{\raggedright Communication interval vs. makespan}
\label{fig6}
\end{figure}

As shown in Fig.~\ref{fig6}, the proposed communication-enabled scheme achieves a 17.27\% scheduling efficiency gain over local reasoning-based scheduling when $D$ is around 25. The optimal interval results from a tradeoff: for small $D$, frequent simultaneous transmissions increase the cyber contention probability, reducing the success rate of route reports $\mathcal{P}_k$, as theoretically analyzed in (\ref{analyze}) and illustrated in Fig.~\ref{fig4}. This communication failure, as shown in Fig.~\ref{fig4}, translates to the poor scheduling performance observed in the left region of Fig.~\ref{fig6}, as AGVs fail to exchange critical route information.
For large $D$, the outdated updates lead to ineffective decisions for AGV scheduling. 
Thus, the optimal interval $D^*$ emerges at the balance point where the rising cyber contention probability for small $D$ intersects with the increasing staleness for large $D$. 
Notably, the optimal communication interval $D^*$ for scheduling in Fig.~\ref{fig6}, differs from the interval setting aimed at maximizing communication throughput in Fig.~\ref{fig4}. 

This distinction indicates that optimizing communication metrics alone in H2H communication does not guarantee optimal performance in M2M communication systems. Thus, an adaptively communication-computation co-design approach is necessary to meet downstream task demands and prioritize overall T-MRS performance.

\begin{figure}[t]
	\centering
\includegraphics[width=0.38\textwidth]{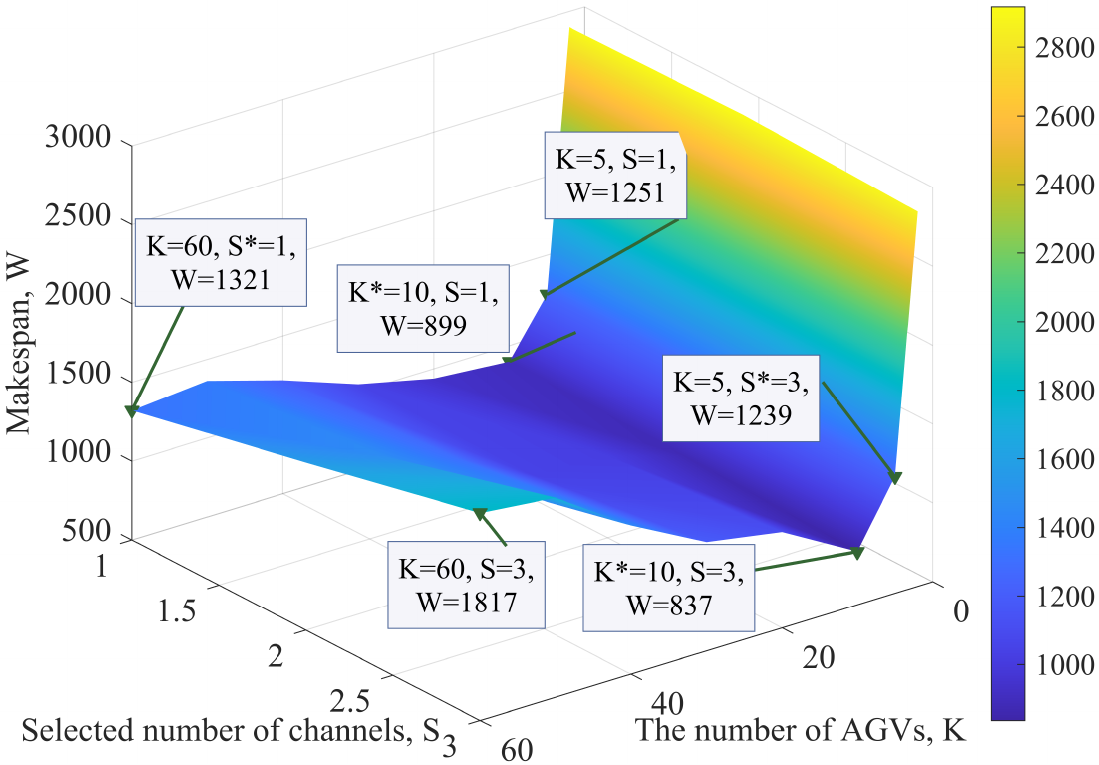}
\caption{\raggedright Makespan $W$ vs. the number of selected channels $S$ and the number of AGVs $K$}
\label{fig7}
\end{figure}

\textbf{The effects of the number of selected channels $S$:} We next investigate the impacts from the number of selected communication channels $S$ in our proposed transmission scheme on the overall scheduling efficiency under varying numbers of AGVs. The experiments are conducted under fixed channel resources ($C=60$), communication interval ($D=2$), and varying numbers of AGVs ($K \in \{2,5,10,20,30,40,50,60\}$). 
Fig. \ref{fig7} shows the overall trend of makespan, which initially decreases and then increases as the number of AGVs grows, regardless of the number of selected channels. This trend is because while more AGVs enable greater production parallelism, they also increase the risk of channel contention, and thus physical congestion and collisions. Fig. \ref{fig7} also demonstrates that an inflection point appears at $K = 10$ when the number of AGVs is 60, beyond which increasing the number of selected channels leads to more channel contention, offsetting the benefits of parallel transmissions. This contention limits AGVs’ ability to receive complete updates of congestion map, thereby worsening partial observability and potentially affecting effective navigation decisions to avoid congestion in advance. Moreover, the optimal number of selected channels for scheduling deviates from that used to maximize communication throughput in Fig.~\ref{fig4}, which suggests the differences between M2M and commonly used H2H communication. While M2M communication prioritizes overall scheduling efficiency in the factory, H2H communication typically aims at maximizing throughput.

These results indicate that multi-link transmission increases transmission reliability and thus improves scheduling efficiency when the ratio of AGVs and available channels is low. As the ratio increases, reservation-based mechanisms provide more effective exchange of route information, supplementing partial observability. Furthermore, it is observed that as the number of AGVs increases, the impact of the number of selected channels on T-MRS scheduling becomes increasingly significant, making adaptive communication strategies essential in larger systems. 

\begin{figure}[t]
	
    \centering
        \includegraphics[width=0.4\textwidth]{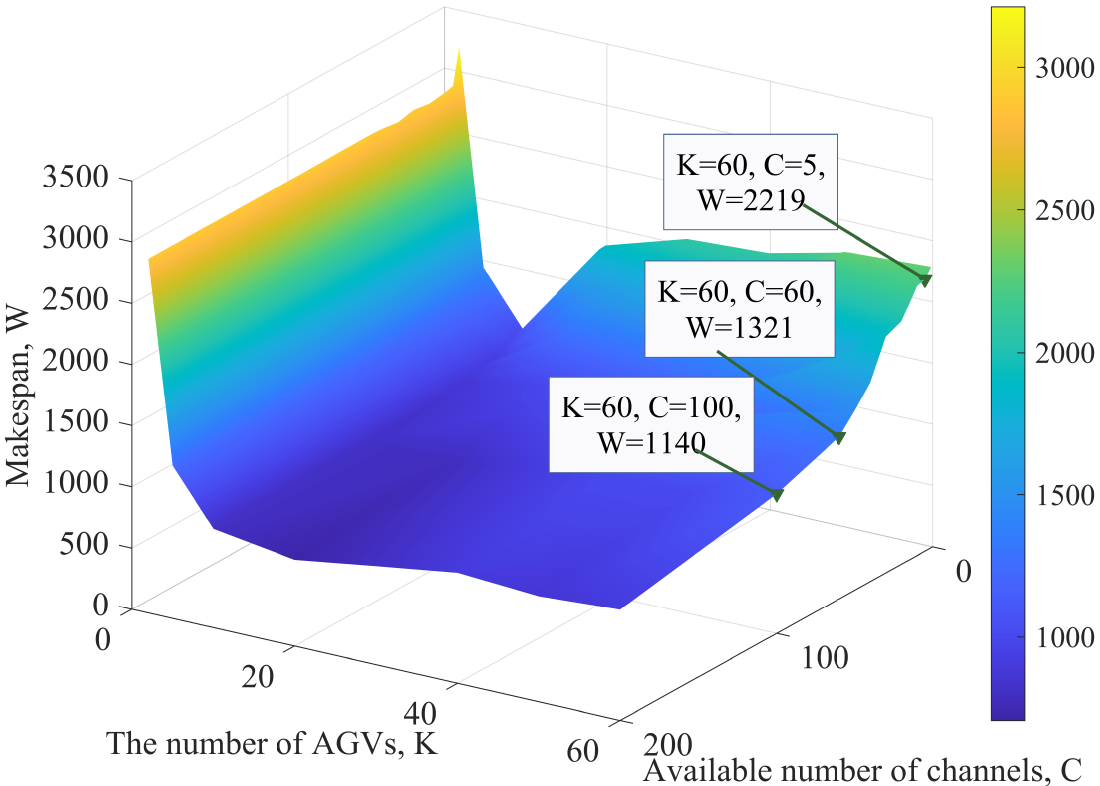}
\caption{\raggedright Makespan $W$ vs. the available number of channels $C$ and the number of AGVs $K$}
\label{fig8}
\end{figure}

\textbf{The effects of the number of available channels $C$:} Finally, we analyze the impact of the number of available channels $C$ on online scheduling using 72 parameter combinations across eight $K$ values and nine $C$ values, as shown in Fig. \ref{fig8}. The results show that increasing the number of available channels consistently enhances information exchange reliability, enabling more AGVs to obtain a reliable global congestion map. This mitigates the limits of partial observability, allowing AGVs to plan effective routes to avoid congestion in advance.
However, a distinct trend emerges: due to the factory size and the number of parallel production lines, the advantages of adding more channel resources $C$ eventually saturate. This finding reveals that co-design of communication and computation is essential, which considers the number of production lines, factory size, the number of AGVs, and the communication channels. 

To sum up, our simulations shows the impacts from wireless communication on the performance of T-MRS and reveal some key insights: 
1) Wireless communication among collaborative AGVs significantly enhances online scheduling efficiency compared to the local reasoning-based baseline, even under high AGV load conditions, see Fig. \ref{fig5}. Meanwhile, even with limited channel resources, the tailored real-time multi-link transmission network effectively supplements partial observability and enhances scheduling efficiency in the proposed dynamic scheduling scheme, see Fig. \ref{fig6}.
2) Moreover, we find that adaptive communication strategies are essential for efficient scheduling performance, see Fig. \ref{fig7} and Fig. \ref{fig8}. Multi-link transmission provides better scheduling efficiency under a low AGV-to-channel ratio, whereas reservation-based access is recommended to enhance scheduling efficiency under a high AGV-to-channel ratio. 
3) Furthermore, the optimal solution for scheduling efficiency in wireless M2M communication differs from the optimal solution in H2H communication, see Fig. \ref{fig4}, Fig. \ref{fig6}, and Fig. \ref{fig7}. This implies a fundamental difference between the M2M and H2H communications and new technological opportunities.

\section{Conclusion}
This study aims to explore how proper information exchange by the tailored wireless M2M networking can supplement partial observability to assist complex computation for online scheduling of T-MRS. To this end, we establish a novel communication-enabled online scheduling system for T-MRS in a smart factory, and then design the scheduling-oriented wireless networking. With the proposed integrated communication and scheduling scheme, AGVs dynamically adjust collision-free and congestion-free routes, significantly enhancing online scheduling efficiency, as demonstrated by extensive results. Beyond performance improvements, the results reveal a fundamental difference between M2M and H2H communication, highlighting the close coupling between communication and computation in T-MRS and offering a fresh perspective for advancing future research in ICC.

Building upon our established online scheduling system, future work will explore two key directions: 1) incorporating distributed multi-agent reinforcement learning for adaptive collision resolution; and 2) employing intelligent radio resource scheduling under uncertainties to further enhance networking reliability and production efficiency. Our ultimate goal is to develop a comprehensive scheduling system that deeply integrates communication, computation, and control for efficient scheduling in real-world smart factories. 

\Acknowledgements{This work was supported by the Joint funds for Regional Innovation and Development of the National Natural Science Foundation of China (No.U21A20449) and the Beijing Natural Science Foundation Program (No.L232002).}

\bibliographystyle{scis}
\small\bibliography{referenceSCIS}

\end{document}